 \documentclass[10pt,twocolumn,a4paper]{article}

 \usepackage{times}
 \usepackage{epsfig}
 \usepackage{graphicx}
 \usepackage{amsmath}
 \usepackage{amssymb}
 \usepackage{multirow} 	 
 \usepackage[super]{nth} 

\newcommand{\argmax}{\operatornamewithlimits{argmax}}

\usepackage{scalerel,stackengine,amsmath}

\usepackage{titling}
\usepackage{authblk}

\usepackage{color, colortbl} 
\definecolor{darkgray}{rgb}{0.7,0.7,0.7}

\usepackage[pagebackref=true,breaklinks=true,colorlinks,bookmarks=false]{hyperref}

\begin{document}

\date{}
\title{Semi-supervised learning of deep metrics for stereo reconstruction}

\author[1]{Stepan Tulyakov \thanks{stepan.tulyakov@epfl.ch}}
\author[1]{Anton Ivanov}
\affil[1]{eSpace Center, Swiss Federal Institute of Technology in Lausanne}
\author[2]{Francois Fleuret}
\affil[2]{Computer Vision and Learning Group, Idiap Research Institute and Swiss Federal Institute of Technology in Lausanne}

\maketitle

\begin{abstract}
Deep-learning metrics have recently demonstrated extremely good performance to match image patches for stereo reconstruction. However, training such metrics requires large amount of labeled stereo images, which can be difficult or costly to collect for certain applications.

The main contribution of our work is a new semi-supervised method for learning deep metrics from unlabeled stereo images, given coarse information about the scenes and the optical system. Our method alternatively optimizes the metric with a standard stochastic gradient descent, and applies stereo constraints to regularize its prediction.

Experiments on reference data-sets show that, for a given network architecture, training with this new method without ground-truth produces a metric with performance as good as state-of-the-art baselines trained with the said ground-truth.

This work has three practical implications. Firstly, it helps to overcome limitations of training sets, in particular noisy ground truth. Secondly it allows to use much more training data during learning. Thirdly, it allows to tune deep metric for a particular stereo system, even if ground truth is not available.
\end{abstract}

\section{Introduction}\label{sec:introduction}

The stereo reconstruction problem consists in estimating a depth map from two images taken from different viewpoints. The problem has many practical applications in robotics~\cite{Menze2015}, remote sensing~\cite{Shean2016}, and 3D graphics~\cite{Strecha2010}.

It has been heavily investigated for several decades~\cite{Scharstein2001}, and recent developments focused on designing high-order, region-based and object-specific priors~\cite{Zhang2015, Chakrabarti2015, Yamaguchi2014, Guney2015, Kim2016, Li2016, Vogel2015, Vogel2014}, and improving efficiency of large scale stereo~\cite{Psota2015, Kowalczuk2012,Geiger2010,Barron2016}. Perhaps the most significant recent breakthrough was to use deep metrics~\cite{Chen2015, Zbontar2015}. It led to considerable gains in processing speed and reconstruction accuracy (see Tables~\ref{tab:benchmarking_mb}, \ref{tab:benchmarking_kitti12}, and ~\ref{tab:benchmarking_kitti15}). Our work improves upon this line of research.

\section{Related work}\label{sec:related_work}

Stereo reconstruction algorithms rely on \textit{epipolar geometry}~\cite{Hartley2003}, according to which to  every no-occluded point in one stereo view corresponds a point in the other view lying on a line that does not depend on the scene, but only on the optical system. This line is called an \textit{epipolar line}, and for a \emph{calibrated} stereo system, it is known for every image point. Furthermore, for a pinhole camera, all the points lying on a given epipolar line in the second view correspond to points lying on a common epipolar line in the first view. Such two epipolar lines are called \textit{conjugate}.

It is a standard procedure to warp stereo views in order to make conjugate epipolar lines in these views horizontal and vertically aligned. This is called \textit{stereo rectification},
and in a rectified stereo pair, every point from the first view corresponds to a point shifted horizontally in the second view. The extension of this shift -- also known as a \textit{disparity} -- allows to compute the distance to the corresponding 3d point, which is the ultimate goal of the stereo reconstruction.

So at the core of the stereo reconstruction process lies the matching of similar patches in two images along epipolar lines and the estimation of the disparity. It is not a trivial task, since the local appearance of a physical point in the two views might differ due to radiometric and geometric distortions. The patch matching is usually performed using \textit{invariant similarity measures} and \textit{descriptors}, also known as features. Historically, the former were more popular for the stereo reconstruction, while the latter were used for matching sparse points of interest.

\subsection{Similarity measures}
The invariant similarity measures~\cite{Hirschmuller2007,Hirschm2008} are popular for stereo reconstruction, probably due to their low computational complexity. The simplest similarity measures are the sum of absolute differences~(SAD), and the sum of squared differences~(SSD). Zero-mean variants of these methods~(ZSAD, ZSSD), as well as sum of absolute gradient differences~(GSAD), are invariant to local brightness changes, which can also be achieved by combining SAD and SSD with background subtraction by mean, Laplacian of Gaussian~(LoG)~\cite{Hirschmuller2002} or Bilateral filters~\cite{Ansar2004}.Non-parametric similarity measures, such as Rank and Census~\cite{Zabih1994} are invariant to arbitrary order-preserving local intensity transformations, and measures such as the Mutual Information~(MI)~\cite{Kim2003} explicitly model the joint intensity distribution in the two images, and are invariant to arbitrary intensity transformations. All these methods are invariant to radiometric distortions only.
\subsection{Descriptors}
Invariant descriptors are popular for sparse point matching, and are designed to be invariant to both radiometric and geometric distortions. They all are either local histograms of oriented image gradients such as SIFT~\cite{Lowe2004}, or binary strings of local pairwise pixel comparisons such as BRIEF~\cite{Calonder2012}. Although descriptors are rarely used for stereo, there are some exceptions, such as DAISY~\cite{Tola2010}, which can be efficiently computed densely.

Recently, the community has moved from these fully hand-crafted descriptors to data-driven descriptors, incorporating machine-learning approaches. Most of such descriptors perform discriminative dimensionality reduction either by feature selection, as VGG~\cite{Simonyan2013}, linear feature extraction, as LDAHash~\cite{Strecha2012}, or boosting, as BinBoost~\cite{Trzcinski2012}.

\subsection{Deep metrics}

As for other application domains of machine learning, the current trend is to move beyond ``shallow'' models, where the learned quantities interact linearly with hand-designed non-linearities, but are not involved in further re-combinations.

The resulting ``deep metrics''
demonstrate extremely good performance compared to other similarity measures and descriptors both for sparse point matching~\cite{Jahrer2008, Fischer2014, Simo-Serra2015, Zagoruko2015, XufengHan2015} and stereo reconstruction~\cite{Zbontar2015, Chen2015}.

Standard deep metric networks have a Siamese architecture, introduced in~\cite{Bromley94}. They consist of two ``embedding'' sub-networks with complete weight sharing that join into a common ``head''. Each embedding sub-network is convolutional, it takes an image patch as input, and outputs the patch's descriptor. The ``head'' is usually fully connected, it takes the two descriptors as input, and outputs a similarity measure. The Siamese architecture was firstly used for image patch matching in its classic form in~\cite{Jahrer2008}. Later it was shown, that the ``head'' network may be replaced by a fixed similarity such as $L^2$~\cite{Simo-Serra2015} or cosine~\cite{Zbontar2015}, that the embedding sub-networks may not share weights~\cite{Zagoruko2015}, and, finally, that the explicit notion of a descriptor might not be necessary~\cite{Zagoruko2015}.

\subsection{Supervised learning of deep metrics}

Existing methods for training a Siamese network for patch matching are supervised, using a training set composed of positive and negative examples. Each positive example (respectively negative) is a pair composed of a reference patch and its matching patch  (respectively a non-matching one) from another image.

Training either takes one example at the time, positive or negative, and adapts the similarity~\cite{Simo-Serra2015, Chen2015, Jahrer2008, Zagoruko2015, XufengHan2015}, or takes at each step both a positive and a negative example, and maximizes the difference between the similarities, hence aiming at making the two patches from the positive pair ``more similar'' than the two patches from the negative pair~\cite{Zbontar2015, Kumar2016, Balntas2016}. This latter scheme is known as ``Triplet Contrastive learning.''

Although the supervised learning of deep metrics works very well, the complexity of the models requires very large labeled training sets which are hard to collect for real applications. Beside, even when such large sets are available, the ground truth is produced automatically from sensors and, thus, usually noisy and/or may suffer from gross  errors.
This can be mitigated by augmenting the training set with random perturbations~\cite{Zbontar2015} or synthetic training data~\cite{Fischer2014,Mayer2016}. However synthesis procedures are hand-crafted and do not account for the regularities specific to the stereo system and target scene at hand.

\subsection{Semi-supervised learning}

Our work is inspired by Multi-Instance Learning (MIL)~\cite{Babenko2008} and Self-Training~\cite{Triguero2013}. The main idea behind MIL, is to use ``coarsely'' labeled data, where one label indicates if a group of samples contains at least one positive sample. This allows to deal with low geometrical accuracy, or even the absence of geometrical information and a labeling at the scene level. It has been applied with success to deep learning~\cite{Wu2015}.

Another strategy to relax the requirement for detailed labeling is Self-Training, where the training set is enriched with unlabeled data. As for transductive learning, self-training works by leveraging the information carried by the unlabeled data about the structure of the data population~\cite{Chen2013, Reed2015}.

Our most efficient method uses dynamic programming (DP) to regularize the noisy prediction of the metric as it is currently trained. Similar idea appeared in~\cite{LeCun1998}, in a different context, to train a deep network to recognize handwritten characters, using word-wise labels to infer character-wise labels. It has also been used to segment automatically sequences of action demonstrations into macro-actions to deal with non-Markovian decision processes~\cite{Lefakis2014}, and the $k$-shortest paths algorithm, which is a generalization of dynamic programming to multiple paths, was used to train a person detector from videos with time-sparse ground-truth~\cite{All2011}.


\section{Method}\label{sec:method}

We start by formulating in \S~\ref{sec:problem} the task of semi-supervised deep metric learning for stereo, then in \S~\ref{sec:constraints} we review the stereo matching problem constraints we consider, and in \S~\ref{sec:methods} we describe how we use them to drive the training.

\subsection{Problem formulation}
\label{sec:problem}

We are provided with a semi-supervised training set $\mathbf{Tr}=\{ (\mathbf{e}^r, \mathbf{e}^+, \mathbf{e}^- )_n \}_{n=1:N} $. Each training example is a triplet of series of $s\times s$ gray-scale patches:
\begin{list}{\labelitemi}{\leftmargin=1.25em}
\itemsep0em
\item \textit{reference patches} $ {\mathbf{e}^{r}=(p^r_1,p^r_2,..,p^r_W)} $ extracted from a horizontal line of a left rectified stereo image,
\item \textit{positive patches} $ {\mathbf{e}^{+}=(p^+_1,p^+_2,..,p^+_W)} $ extracted from the corresponding horizontal in the right rectified stereo image, and
\item \textit{negative patches} $ {\mathbf{e}^{-}=(p^-_1,p^-_2,..,p^-_W)} $ extracted from another horizontal line of a right rectified stereo image,
\end{list}
where $ W $ is the number of patches per line, and $N$ is the number of training examples. In addition to the training set, we are provided with the maximum possible disparity $ d_{max} $, which depends on the optical system and a prior knowledge about the scene.

Our goal is to learn a deep metric $S(x, y)$ such that, for any set of reference $\mathbf{e}^r$ and positive image patches $\mathbf{e}^+$, the row-wise maxima of the \textit{similarity matrix} $ \mathbf{S}^{r+}_{ij}=S\left(p^r_i, \, p^+_j\right)$ correspond to the true matches.

Note, that in contrast to~\cite{Jahrer2008,Zagoruko2015,Simo-Serra2015,Fischer2014,XufengHan2015,Chen2015,Zbontar2015} in our case each training example is not a pair of patches, but a triplet of series of patches each taken on an horizontal line of a rectified stereo image, so that we can utilize constraints and loss functions defined on such families of patches jointly. Additionally, processing lines as a whole significantly speeds up the training process by allowing to reuse shared computations.

\subsection{Matching constraints}
\label{sec:constraints}

The stereo matching problem satisfies the following constraints:
\begin{list}{\labelitemi}{\leftmargin=1.25em}
\itemsep0em

\item[(E)] \textbf{Epipolar constraint.} Every non-occluded reference patch has a matching positive patch~\cite{Hartley2003}[239-241p].

\item[(D)] \textbf{Disparity range constraint.} The offset of the reference patch index with respect to the matching positive patch index is bounded by a maximum disparity $ d_{max} $. This comes from the stereo system parameters (focal length, pixel size, baseline) and the distance range of the scenes.

\item[(U)] \textbf{Uniqueness constraint.} The matching positive patch is unique~\cite{Marr1979}.

\item[(C)] \textbf{Continuity constraint.} The offsets of the reference patches indices with respect to the matching positive patch indices are similar for nearby reference patches everywhere except on depth discontinuities~\cite{Marr1979}.

\item[(O)] \textbf{Ordering constraint.} The reference patches are ordered on their lines as the matching positive patches on theirs.
\end{list}

\begin{figure}[hbtp]
\centering
\begin{minipage}{.5\textwidth}
  \centering
  \includegraphics[width=0.95\textwidth]{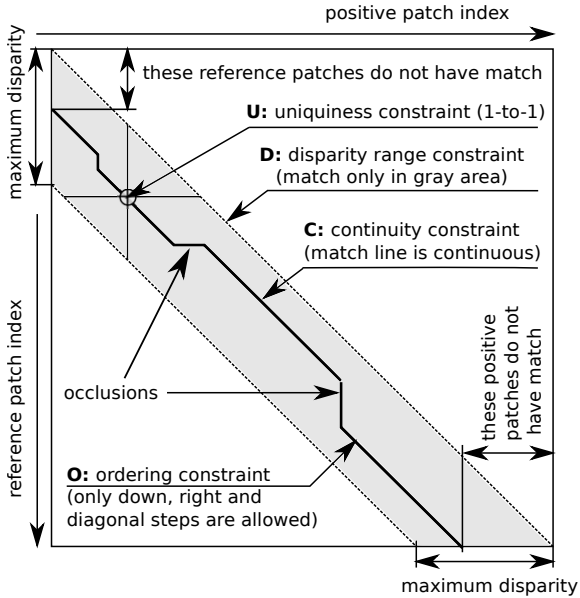}
  \caption{Positive similarity matrix. The bold line corresponds to the optimal matches that satisfy the stereo constraints. Elements within the disparity range are shown in gray. Note that there are no matches for some points on the reference and positive epipolar lines.}
 \label{fig:constraints}
\end{minipage}%
\end{figure}

These constraints result in a particular shape of the positive similarity matrix, as pictured in Figure~\ref{fig:constraints}.

\subsection{Proposed semi-supervised methods}
\label{sec:methods}

We developed several semi-supervised methods that use different subsets of the stereo constraints during training. All methods alternate between two steps: (1) improving the metric, given the current estimate of the matches for the positive examples, and (2) re-computing these matches under the constraints, given the current estimate of the metric. They can be used in combination with any deep metric architecture and any gradient based optimization method.

To each of our methods corresponds a loss function optimized in each of the two steps mentioned above. It takes as an input either $\mathbf{S}^{r+}$, or the three matrices  $\mathbf{S}^{r+}$, $\mathbf{S}^{r-}$ and $\mathbf{S}^{-+}$ defined respectively as follows:
\begin{align}
S^{r+}_{ij} & = \left\{ \begin{array}{ll} S(p_i^r, p_j^+) & 0\le i-j\le d_{max} \\ -\infty & otherwise \\ \end{array} \right.\\
S^{r-}_{ij} & = \left\{ \begin{array}{ll} S(p_i^r, p_j^-) & 0\le i-j\le d_{max} \\ -\infty & otherwise \\ \end{array} \right.\\
S^{-+}_{ij} & = \left\{ \begin{array}{ll} S(p_i^-, p_j^+) & 0\le i-j\le d_{max} \\ -\infty & otherwise \\ \end{array} \right.
\end{align}
In the next sections we describe each method in details.

\subsubsection{MIL method}\label{sec:mil}

This method is inspired by Multi-Instance Learning (MIL) paradigm~\cite{Babenko2008} and uses only the epipolar and the disparity range constraints (E) and (D) from \S~\ref{sec:constraints}.

From these two constraints, we know that every non-occluded reference patch has a matching positive patch in a known index interval, but does not have a matching negative patch. Therefore, for every reference patch, the similarity of the best reference-positive match should be greater than similarity of the best reference-negative match. Our training objective is to push apart these two similarities.

The training loss for the MIL method is
%
%
\begin{multline}\label{eq:mil}
\!\!\!\! \frac{1}{|\mathbf{rows}|} \sum_{i \in \mathbf{rows}} \max(0, -\max_jS^{r+}_{ij} + \max_jS^{r-}_{ij} + \mu) \ \ + \\
\frac{1}{|\mathbf{cols}|} \sum_{j \in \mathbf{cols}}\max(0, -\max_iS^{r+}_{ij} + \max_iS^{-+}_{ij} + \mu),
\end{multline}
where
%
$ \mathbf{rows} = \{d_{max}+1, \dots, W \} $ is a set of  rows of the similarity matrix that are guaranteed to have correct matches (see Fig~\ref{fig:constraints}), $ \mathbf{col} = \{1, \dots, W-d_{max} \} $ is a set of valid columns of the similarity matrix that are guaranteed to have correct matches, $ W $ is the number of patches in a horizontal line of rectified image, and $ \mu $ is a loss margin. Note that the disparity range constraint is taken into account automatically, if we use the similarity matrices as defined in \S~\ref{sec:methods}.

Experiments shows that the method learns metrics insensitive to small shifts from the optimal match. This problem results in blocky shape of a similarity matrix, where blocks correspond to the areas where the metric is not able to find unique match. This issue motivates the CONTRASTIVE method described in the following section.

\subsubsection{CONTRASTIVE method}
\label{sec:contrastive}

This method uses the epipolar, the disparity range, and the uniqueness constraints (E), (D), and (U) from \S~\ref{sec:constraints}.

From the epipolar and the disparity range constraints we know that every non-occluded reference patch has a matching positive patch in a known index interval. Furthermore, according to the uniqueness constraint the matching positive patch is unique. Therefore, for every patch, the similarity of the best match should be greater than the similarity of the second best match. Our training objective is to push apart these two quantities.

The training loss for this CONTRASTIVE method is
\begin{multline}\label{eq:contrastive}
\!\!\!\! \frac{1}{|\mathbf{rows}|} \sum_{i \in \mathbf{rows}} \max(0, -\max_jS^{r+}_{ij} + \max_{j}\hat{S}^{r+}_{ij} + \mu) \ \ + \\
\frac{1}{|\mathbf{cols}|} \sum_{j \in \mathbf{cols}}\max(0, -\max_iS^{r+}_{ij} + \max_i {\check{S}}^{r+}_{ij} + \mu),
\end{multline}
where $ \hat{\mathbf{S}} $ is a similarity matrix with masked out row-wise maxima, $ {\check{\mathbf{S}}} $ is a similarity matrix with masked out column-wise maxima. To mask out elements of similarity of matrix, we simply substitute them with $ -\infty $.

Experiments show that this method suffers from a problem opposite to the one exhibited by the MIL method: it produces over-sharpened metric, sensitive even to small shifts from the exact match. This is also detrimental to the performance, since our goal is to find metric invariant to small geometric transformations, such as shift. We solved the problem by masking out all spatial neighbors withing $ t_{sup} $ radius from the maximas in $ \hat{\mathbf{S}} $ and in $ {\check{\mathbf{S}}} $. See the supplementary materials for details.

\subsubsection{MIL-CONTRASTIVE method}
\label{sec:mil_contrastive}

As we showed in previous sections, the CONTRASTIVE and the MIL methods have complementary properties and use the stereo constraints in orthogonal way. Therefore we can combine them into a new method that we call MIL-CONTRASTIVE.

\subsubsection{CONTRASTIVE-DP method}
\label{sec:contrastive_dp}

This method uses all constraints listed in \S~\ref{sec:constraints}. The only difference with CONTRASTIVE is that it finds the best match under (C) and (O) using dynamic programming (DP), instead of independent maxima.

Formally, it solves
\begin{equation}\label{eq:dprog_objective}
p^* = \argmax_{p \in {\mathcal P}} \frac{1}{|p|}\sum_{(i, j) \in p } \mathbf{S}^{r+}_{ij}, \\
\end{equation}
where ${\mathcal P}$ is the set of paths $\{ (i_n, j_n) \}_{n=1:M}$
which are continuous in the following sense:
%
\begin{align*}
&\forall n\!>\!1, (i_n, j_n) - (i_{n-1},j_{n-1}) \in \{ (0,1), (1,0), (1,1) \}, \\
&\text{and} \ \ (i_1, j_1) \in \{ 1 \} \times [1, d_{max}].
\end{align*}
Which means that only down, right and diagonal steps are allowed. This enforces the continuity and the ordering constraints (C) and (O) in the solution. Notice also that we search for a path that has maximum average energy rather than maximum total energy to prevent a bias toward longer paths and consequently smaller disparities.

Given the best match-path $p^*$ found by the dynamic programming we define our loss function as
\begin{align}\label{eq:contrastive_dp}
& \frac{1}{|p^*|} \sum_{(i,j) \in p^*} \max(0, -S^{r+}_{ij} + \max_k\tilde{S}^{r+}_{ik} + \mu)+ \nonumber \\ 
& \frac{1}{|p^*|} \sum_{(i,j) \in p^*}\max(0, -S^{r+}_{ij} + \max_l\tilde{S}^{r+}_{lj} + \mu),
\end{align}
where $ \tilde{\mathbf{S}} $ is a similarity matrix where all neighbors of elements belonging to $ p^* $ withing radius $ t_{sup} $ are masked out by setting their values to $-\infty$.

The best match-path computed by the dynamic programming might contain vertical and horizontal segments. These segments correspond to patches that are occluded by foreground objects on one of the views, and thus do not have correct matches. Therefore, in our experiments we ignore vertical and horizontal segments longer than $ t_{occ} $ during the learning. For more details, please refer to the supplementary materials.

\section{Experiments}

Our experiments were done in the Torch framework~\cite{Collobert2011}. Optimization was performed with the ADAM method with standard settings, using mini-batches of size equal to the training images height, and no data augmentation of any sort. The initialization of weights and biases of our deep metric network was done in standard way by random sampling from zero-mean uniform distribution.  

We guarantee reproducibility of all experiments in this section by using only available data-sets, and making our code available online under open-source license after publications.

\subsection{Data-Sets}

In our experiments we use three popular benchmark data-sets: KITTI'12~\cite{Geiger2012}, KITTI'15~\cite{Menze2015} and Middlebury~(MB)~\cite{Scharstein2001,Scharstein2003, Scharstein2007, Hirschmuller2007, Scharstein2014}. These data-sets have online scoreboards~\cite{KITTI, MB}, showing comparative performance of all participating stereo methods.


KITTI'12 and KITTI'15 data-sets each consist of 200 training and 200 test rectified stereo pairs of resolution 1226$\times$370 acquired from cars moving around a city. About $30\%$ of the pixels in the training set are supplied with a ground truth disparity acquired by a laser altimeter with error less than 3 pixels. The disparity range is about 230 pixels. Each data-set is supplied with an extension (respectively KITTI'12-EXT and KITTI'15-EXT) that contains 19 additional stereo pairs for each scene, without ground truth disparity. This allows us to use 40$\times$ more training data for the semi-supervised learning than for the supervised (actually even more, considering that only about 30\% of pixels in the training set have labels).

Middlebury data-set (MB) consists of 60 training and 30 test rectified stereo pairs. The images are acquired by different stereo systems and contain different artificial scenes. Their resolution varies from 380$\times$430 to 3000$\times$2000, and their disparity ranges vary from 30 to 800 pixels. The training images are provided with a dense ground truth disparity acquired by structured light system with error less that 0.2 pixels.

\subsection{Performance measure}\label{sec:performance-measure}

To estimate the performance of deep metrics we compute a prediction error rate defined as the proportion of non-occluded patches for which the predicted disparity is off by more than 3 pixels.

The motivation behind this work is to improve the metric as a mean to match patches in a stand-alone manner, as we have not taken into account the interplay with the additional post-processing that may be applied in a complete stereo pipeline. Performance regarding this main objective is measured by picking the patch with the largest similarity among the patches that belong to a valid disparity range on the epipolar line. We call this the \emph{winner-take all} (WTA) error rate.

A second measure is the error rate of a complete stereo pipeline with plugged-in deep metric. This is a performance measure of direct practical interest, although not the objective we optimize during our training.

\subsection{Deep metric architecture}

The main contribution of this work is a new semi-supervised training method, not deep metric architecture, therefore we simply adopt the overall architecture of well performing MC-CNN~fst network from~\cite{Zbontar2015}, shown in Table~\ref{tab:metric_architecture}, and substitute their learning method with ours.  
\begin{table}[ht!]
\resizebox{\columnwidth}{!}{%
\begin{tabular}{lll}
\hline
 \textbf{Parameter}           & \textbf{KITTI'12,15} & \textbf{MB} \\
\hline
 Number of CNN layers         & 4                    & 5                         \\
 Number of features per layer & 64                   & 64                        \\
 Receptive field              & 3x3x64               & 3x3x64                    \\
 Activation function          & ReLU                 & ReLU                      \\
 Equivalent patch size        & 9x9                  & 11x11                     \\
 Similarity metric            & Cosine               & Cosine                    \\
\hline
\end{tabular}
}
\vspace{1mm}
\caption{Network architectures for deep metric from~\cite{Zbontar2015} that we use in our experiments.}\label{tab:metric_architecture}
\end{table}

\subsection{Comparison of semi-supervised methods}
\label{sec:compare_semisupervised_method}

In this experiment we compare the performance of the proposed semi-supervised methods. We performed comparison on KITTI'12 data-set using the winner-take-all (WTA) error (see \S~\ref{sec:performance-measure}). The results of the experiments are shown in Table~\ref{tab:compare_semisupervised_methods}.

\begin{table}[ht!]
\resizebox{\columnwidth}{!}{%
\begin{tabular}{lll}
\hline
\textbf{Method} & \textbf{WTA error, [$\%$]} & \textbf{Time, [hr]} \\
\hline
MIL 	   			 	    			& 18.45  & 45  \\
CONTRASTIVE        						& 17.63  & 30  \\
MIL-CONTRASTIVE    						& 16.12  & 65  \\
\textbf{CONTRASTIVE-DP} 				& \textbf{14.61}  & \textbf{68} \\
\hline
\end{tabular}
}
\vspace{1mm}
\caption{Comparison of the proposed semi-supervised learning methods on KITTI'12 set. All methods are used to train the same network architecture. The CONTRASTIVE-DP method, shown in bold, uses all the constraints during learning and achieves the smallest WTA error. Notice that in general increasing the number of constraints increases performance.}\label{tab:compare_semisupervised_methods}
\end{table}

The main conclusion is that semi-supervised methods that use more stereo constraints during learning perform better. For example, the MIL, that uses only the epipolar and the disparity range constraints, has larges WTA error, whereas the CONTRASTIVE-DP, that uses the epipolar, the disparity range, the continuity, the uniqueness and the ordering constraints has smallest WTA error.

In all following sections, we use the best performing CONTRASTIVE-DP method only, and refer to it as MC-CNN-SS, where SS stands for semi-supervised.

\subsection{Comparison with supervised method}\label{sec:comp-with-superv}

In this section, we compare the proposed semi-supervised method with our reference fully supervised deep-metric baseline~\cite{Zbontar2015} on the three different sets, using the winner-take-all (WTA) error (see \S~\ref{sec:performance-measure}).

The results are shown in Table~\ref{tab:comparision_to_supervised}. As we see, our method outperforms the supervised method in terms of WTA error across two sets, and does virtually as well on the third. This is remarkable considering the fact that our method does not use ground truth disparity during learning.

The success of our method in case of KITTI'12 and KITTI'15 sets can be attributed to the fact that these sets have large amount of unlabeled stereo data, that can be used by our method. In fact, these sets have more than $40\times$ more unlabeled data than labeled training data.

In case of MB data-set our method does not have this advantage over the supervised method. The set has only 30\% more unlabeled training data than the labeled training data. This is probably the reason why our method shows slightly worse performance on this dataset than compared to the supervised method.

\begin{table}[ht!]
\resizebox{\columnwidth}{!}{%
\begin{tabular}{llll}
\hline
\multirow{2}{*}{\textbf{Method}}& \multicolumn{3}{c}{\textbf{WTA error, [\%]}} \\
								& \textbf{KITTI'12} & \textbf{KITTI'15} & \textbf{MB}\\
\hline
MC-CNN~fst~\cite{Zbontar2015}	& 15.44 & 15.38 & \textbf{29.94}  \\
MC-CNN-SS~fst (ours)  & \textbf{13.90} & \textbf{14.08} & 30.06 \\
\hline
CENSUS~9x9~\cite{Zabih1994}     & 53.52 & 50.35 & 64.53 \\
AD~9x9            				& 32.36 & 30.67 & 59.39 \\
\hline
\end{tabular}
}
\caption{Comparison of our semi-supervised learning method with the fully supervised baseline using the same network architecture~\cite{Zbontar2015}. Smallest WTA errors are shown in bold. Our semi-supervised method outperforms the baseline in terms of WTA error across two sets, and does virtually as well on the third. This is remarkable since in contrast to the supervised method, our does not use ground truth disparity during learning. For reference, the two bottom rows show the performance of two standard similarity measures and descriptors. Note that following the setup of~\cite{Zbontar2015}, the patches used as input to the deep-learning methods are of size $9 \times 9$ for KITTI'12,'15, and $11 \times 11$ for MB.}\label{tab:comparision_to_supervised}
\end{table}

\subsection{Stereo benchmarking}

In this section we investigate how well our semi-supervised deep metric performs when it is combined with the complete stereo pipeline. For that we plug it in the stereo pipeline from~\cite{Zbontar2015}, and tuned the parameters of the pipeline using simple coordinate descent method, starting from the default values of~\cite{Zbontar2015}. Note that we used specific metric and pipeline parameters for each data-set.

Then we computed disparity maps for the test sets with withheld ground truth, and uploaded the results to the evaluation web sites for the respective data-sets\cite{KITTI,MB}. The obtained evaluation results are shown in Tables \ref{tab:benchmarking_kitti12}, \ref{tab:benchmarking_kitti15} and \ref{tab:benchmarking_mb}. As we can see, results with our metric trained without ground truth during training are very close to the results of the fully supervised method across all benchmarks.

Those are very encouraging results, given in particular that we did not optimize the deep metric and the pipeline parameters together, and considering the performance in the winner-take-all setup of \S~\ref{sec:comp-with-superv}.

Regarding the processing time, note that the network structure used for our method is identical to that of MC-CNN-fst~\cite{Zbontar2015}, except for the pipeline parameters. The difference in processing times in Tables \ref{tab:benchmarking_kitti12}, \ref{tab:benchmarking_kitti15} and \ref{tab:benchmarking_mb} is only due to the hardware differences.
\begin{table}[ht!]
\resizebox{\columnwidth}{!}{%
\begin{tabular}{ m{0.01\columnwidth} m{0.13\columnwidth} m{0.37\columnwidth}  m{0.33\columnwidth}   m{0.16\columnwidth} }
\hline
\textbf{\#} & \textbf{Date} & \textbf{Algorithm} & \textbf{Pipeline Err, [\%]} & \textbf{Time, [s]}\\
\hline
1 & 01/19/15 & NTDE~\cite{Kim2016} & 7.62 & 300 \\
2 & 08/28/15 & MC-CNN~acrt~\cite{Zbontar2015} & 8.29 & 254 \\
3 & 11/03/15 & MC-CNN+RBS ~\cite{Barron2016} & 8.62 & 345 \\
\textbf{4} & \textbf{01/26/16} & \textbf{MC-CNN~fst~\cite{Zbontar2015}} & \textbf{9.69} & \textbf{2.94} \\
\textbf{5} & \textbf{14/11/16} & \textbf{MC-CNN-SS~(ours)} & \textbf{12.3} & \textbf{5.59}\\
6 & 10/13/15 & MDP~\cite{Li2016} & 12.6 & 130 \\
7 & 04/19/15 & MeshStereo~\cite{Zhang2015} & 13.4 & 146  \\
\hline
\end{tabular}
}
\vspace{1mm}
\caption{MB benchmark~\cite{MB} snapshot from 14/11/2016 with published methods (default view). Methods ranked 1, 2, 3, 4 and 5 use deep metrics for stereo matching. Note that our semi-supervised method MC-CNN-SS, shown in bold, that does not use ground truth data during training, has an error rate very similar to that of the supervised MC-CNN~fst method, also shown in bold, trained with ground truth data. }\label{tab:benchmarking_mb}
\end{table}

\begin{table}[ht!]
\resizebox{\columnwidth}{!}{%
\begin{tabular}{  m{0.01\columnwidth} m{0.13\columnwidth} m{0.37\columnwidth}  m{0.33\columnwidth}   m{0.16\columnwidth} }
\hline
\textbf{\#} & \textbf{Date} & \textbf{Algorithm} & \textbf{Pipeline Err, [\%]} & \textbf{Time, [s]}\\
\hline
1 & 27/04/16 & PBCP~\cite{Seki2016} & 2.36 & 68 \\
2 & 26/10/15 & Displets v2~\cite{Guney2015} & 2.37 & 265 \\
3 & 21/08/15 & MC-CNN acrt~\cite{Zbontar2015} & 2.43 & 67 \\
4 & 30/03/16 & cfusion~\cite{Ntouskos2016} & 2.46 & 70 \\
5 & 16/04/15 & PRSM~\cite{Vogel2015} & 2.78 & 300 \\
\textbf{6} & \textbf{21/08/15} & \textbf{MC-CNN fst~\cite{Zbontar2015}} &\textbf{2.82} & \textbf{0.8} \\
7 & 03/08/15 & SPS-st~\cite{Yamaguchi2014} & 2.83 & 2  \\
\textbf{8} & \textbf{14/11/16} & \textbf{MC-CNN-SS~(ours)} & \textbf{3.02} & \textbf{1.35}\\
9 & 03/03/14 & VC-SF~\cite{Vogel2014} & 3.05 & 300 \\
\hline
\end{tabular}
}
\vspace{1mm}
\caption{KITTI'12 benchmark~\cite{KITTI} snapshot from 14/11/2016 with published  methods (default view). Methods ranked 1, 2, 3, 4, 6, and 7 use deep metrics for stereo matching. Note that our semi-supervised method MC-CNN-SS, shown in bold, that does not use ground truth data during training, has an error rate very similar to that of the supervised MC-CNN~fst method, also shown in bold, trained with ground truth data. Since the MC-CNN~fst method does not appear on KITTI'12 evaluation table, due to restrictions on the number of results for a single paper, we borrowed it from ~\cite{Zbontar2016}}\label{tab:benchmarking_kitti12}
\end{table}

\begin{table}[ht!]
\resizebox{\columnwidth}{!}{%
\begin{tabular}{   m{0.01\columnwidth} m{0.13\columnwidth} m{0.37\columnwidth}  m{0.33\columnwidth}   m{0.16\columnwidth}  }
\hline
\textbf{\#} & \textbf{Date} & \textbf{Algorithm} & \textbf{Pipeline Err, [\%]} & \textbf{Time, [s]}\\
\hline
1 & 26/10/15 & Displets v2~\cite{Guney2015} & 3.43 & 265 \\
2 & 27/04/16 & PBCP~\cite{Seki2016} & 3.61 & 68 \\
3 & 21/08/15 & MC-CNN~acrt~\cite{Zbontar2015} & 3.89 & 2.94 \\
4 & 16/04/15 & PRSM~\cite{Vogel2015} & 4.27 & 300 \\
5 & 06/11/15 & DispNetC~\cite{Mayer2016}  & 4.34 & 0.06 \\
6 & 11/04/16 & ContentCNN~\cite{Luo2016}  & 4.54 & 1 \\
\textbf{7} & \textbf{21/08/15} & \textbf{MC-CNN~fst~\cite{Zbontar2015}} & \textbf{4.62} & \textbf{0.8} \\
\textbf{8} & \textbf{14/11/16} & \textbf{MC-CNN-SS~(ours)} & \textbf{4.97} & \textbf{1.35} \\
9 & 03/08/15 & SPS-st~\cite{Yamaguchi2014} & 5.31 & 2 \\
\hline
\end{tabular}
}
\vspace{1mm}
\caption{KITTI'15 benchmark~\cite{KITTI} snapshot from 14/11/2016 with published methods (default view). Methods ranked 1, 2, 3, 5, 6 and 7 use deep metrics for stereo matching. Note that our semi-supervised method MC-CNN-SS, shown in bold, that does not use ground truth data during training, has an error rate very similar to that of the supervised MC-CNN~fst method, also shown in bold, trained with ground truth data. Since the MC-CNN~fst method does not appear on KITTI'12 evaluation table, due to restrictions on the number of results for a single paper, we borrowed it from ~\cite{Zbontar2016}.}\label{tab:benchmarking_kitti15}
\end{table}

\subsection{What does deep metric learn?}
\label{sec:what_is_learned}

\begin{figure}[hbtp]
\centering
\begin{minipage}{.5\textwidth}
  \centering
  \includegraphics[width=0.95\textwidth]{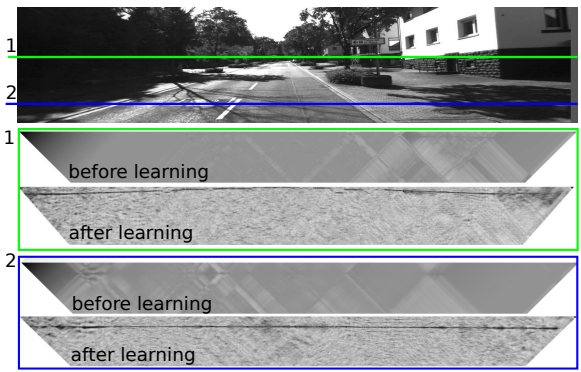}
  \caption{Diagonal part of the similarity matrix before and after training with MC-CNN-SS on KITTI'12 dataset. Top figure shows one of the stereo images with two highlighted epipolar lines. The pictures below show the positive similarity matrices for these epipolar lines. The dark elements in the similarity matrices correspond to the higher similarities. WTA error before training is 42.01\%, and 14.61\% after. Note that before the training we can not visually distinguish the best matches in the similarity matrices, while after the learning they are clearly visible.}
 \label{fig:similarity_matrix}
\end{minipage}%
\end{figure}

In Figure~\ref{fig:similarity_matrix} we show positive similarity matrices before and after the training with MC-CNN-SS on KITTI'12 data-set. While one can not visually distinguish the best match in the similarity matrices before the training, it becomes clearly visible after. This suggests that the training improves discriminative ability of the deep metric.

\begin{figure}[hbtp]
\centering
\begin{minipage}{.5\textwidth}
  \center
  \includegraphics[width=\textwidth]{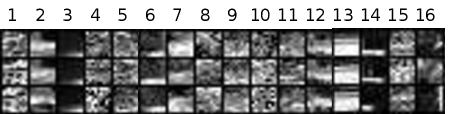}
  \caption{Failure cases of the deep metric trained with our MC-CNN-SS method on the KITTI data-set. For each example the three patches displayed correspond to (from top to bottom): the reference patch, the predicted match and the ground-truth match. Note that as expected, the ground truth and the predicted matches are often visually indistinguishable. This happens if the reference patch is from an area with almost horizontal edges (3, 6, 13), a flat image area (4, 5, 10), or an area with repetitive texture. Some failures are triggered by likely errors in the ground truth labeling (2, 12, 14, 16).}
\label{fig:failure_cases}
\end{minipage}%
\end{figure}

In Figure~\ref{fig:failure_cases} we show failure cases of learned deep metric. Most of the failures happen when the ground truth match is visually indistinguishable from the incorrect match picked by the deep metric. This happens if the reference patch is from a flat image area, an area with a repetitive texture, or an area with a horizontal edge.

Notably, some failures are triggered by probable errors in the ground truth. These errors might worsen outcomes of the supervised learning but does not affect outcomes of our semi-supervised learning, since it does not use the ground-truth.





\subsection{Generalization across data-sets}

In this experiment, we study how deep metric trained using our semi-supervised method on one data-set performs on another data-sets in terms of WTA error.

From Table~\ref{tab:generalization} it appears that a metric always performs better when the train and test population come from the same data-set. This confirms that our semi-supervised metric has great practical value: it allows to tune descriptor for a particular stereo system at hands, even if data-set with ground-truth is not available.

\begin{table}[ht!]
\center
\begin{tabular}{llll}
\hline
\multirow{2}{*}{\textbf{Training set}}& \multicolumn{3}{c}{\textbf{WTA error, [\%]}} \\
								& \textbf{KITTI'12} & \textbf{KITTI'15} & \textbf{MB}\\
\hline
KITTI'12    & \textbf{13.90} & 15.52 		  & 34.85   \\
KITTI'15    & 16.61 		 & \textbf{14.08} & 36.66  \\
MB  & 14.22 		 & 15.00 		  & \textbf{30.06}   \\
\hline
\end{tabular}
\vspace{1mm}
\caption{Generalization error across data-sets. The smallest WTA errors, shown in bold correspond to the cases when the train and test population come from the same data-set. This confirms that our semi-supervised metric has great practical value: it tune the descriptors for a particular stereo system, even if no ground truth is available.}\label{tab:generalization}
\end{table}





\section{Conclusion}

We proposed novel semi-supervised techniques for training patch similarity measures for stereo reconstruction. These techniques allow to train with data-sets for which ground truth is not available, by relying on simple constraints coming from properties of the optical sensor, and from a rough knowledge about the scenes to process.

We applied this framework to the training of a ``deep metric'', that is a deep siamese neural-network that takes two patches as an input and predicts a similarity measure. Benchmarking on standard data-sets shows that the resulting performance is as good or better than published results with the same network trained on the same but fully labeled data-sets (see Table~\ref{tab:comparision_to_supervised}).

This very good performance can be explained by the strong redundancy of a fully labeled data-set, due to the continuity of surfaces, coupled with inevitable labeling errors. The latter can degrade the performance resulting from a fully supervised training process, and could only be mitigated by using a prior knowledge about the regularity of the labeling, similar to the constraints we use.

The techniques we propose open the way first to using stereo reconstruction based on deep metrics for data-sets for which no ground-truth exists, such as planetary measurements. Second, it will allow the training of larger neural networks, with very large unlabeled data-sets. Our experiments show that the network that we are using in our experiments does benefit from an one order of magnitude more training samples, than it is available to supervised method as shown in Table \ref{tab:comparision_to_supervised}. We expect that this effect will be even more significant if we use our training method with larger networks that would over-fit existing labeled training sets.

{\small
\bibliographystyle{ieee}
\bibliography{cnnmars}
}

\end{document}